# MILP, Pseudo-Boolean, and OMT Solvers for Optimal Fault-Tolerant Placements of Relay Nodes in Mission Critical Wireless Networks* †

**Qian Matteo Chen**
*Computer Science Department, Sapienza University of Rome, Italy*

**Alberto Finzi**
*Department of Electrical Engineering and Information Technology, University of Naples Federico II, Italy*

**Toni Mancini** C
*Computer Science Department, Sapienza University of Rome, Italy*

**Igor Melatti**
*Computer Science Department, Sapienza University of Rome, Italy*

**Enrico Tronci**
*Computer Science Department, Sapienza University of Rome, Italy*

**Abstract.** In *critical infrastructures* like airports, much care has to be devoted in protecting radio communication networks from external electromagnetic interference.

Protection of such *mission-critical* radio communication networks is usually tackled by exploiting *radiogoniometers*: at least three suitably deployed radiogoniometers, and a gateway gathering information from them, permit to monitor and localise sources of electromagnetic emissions that are not supposed to be present in the monitored area. Typically, radiogoniometers are connected to the gateway through *relay nodes*. As a result, some degree of fault-tolerance for the network of relay nodes is essential in order to offer a reliable monitoring. On the other hand, deployment of relay nodes is typically quite expensive. As a result, we have two conflicting requirements: minimise costs while guaranteeing a given fault-tolerance.

In this paper, we address the problem of computing a deployment for relay nodes that minimises the relay node network cost while at the same time guaranteeing proper working of the network even when some of the relay nodes (up to a given maximum number) become faulty (*fault-tolerance*).

We show that, by means of a computation-intensive pre-processing on a HPC infrastructure, the above optimisation problem can be encoded as a 0/1 Linear Program, becoming suitable to be approached with standard Artificial Intelligence reasoners like MILP, PB-SAT, and SMT/OMT solvers. Our problem formulation enables us to present experimental results comparing the performance of these three solving technologies on a real case study of a relay node network deployment in areas of the Leonardo da Vinci Airport in Rome, Italy.

# Contents

# 1.  Introduction

The operations of several *critical infrastructures* (see, *e.g.*, [2]) like airports rely on the proper functioning of radio communication networks [3]. Such communication networks are indeed *mission-critical communication networks* (see, *e.g.*, [4]), and require adequate protection from external electromagnetic interferences.

For instance, in airports, *ground movement control* is responsible for areas such as: taxiways, inactive runways, holding areas, and some transitional aprons or intersections where aircrafts arrive, having vacated runways or departure gates. In this context, communications between the air traffic control tower and vehicles, or personnel moving in the airport area takes place through VHF/UHF radio (see, *e.g.*, [5]). In these cases, an electromagnetic attack (*e.g.*, through *jamming*) preventing communication between the air traffic control tower and aircraft would degrade the airport operations, and could possibly yield serious safety threats (*e.g.*, [6, 7, 8, 9]).

Accordingly, the area where the mission-critical communication network is deployed is typically *monitored* in order to *detect* spurious electromagnetic emissions and *localise* the source of such emissions (*transmitter*). This is usually done exploiting *radiogoniometers*: at least three suitably deployed radiogoniometers, along with a gateway gathering information from them, enable to monitor and localise sources of electromagnetic emissions that are not supposed to be present in the monitored area.

## 1.1.  Motivation

Unfortunately, effective deployment of radiogoniometers in such environments is all but easy. In fact, communication between the radiogoniometers and the gateway must use a spectrum that does not interfere with the one used within the infrastructure under protection. For this purpose, often communication between radiogoniometers and the gateway rests on an unlicensed spectrum such as WiFi.

This has far-reaching implications. The limitations on the transmission power for WiFi channels, for a large area like a big airport, hamper single hop communications between the radiogoniometers and the gateway. Hence, *relay* nodes need to be deployed in the area to ensure that radiogoniometers can always communicate with the gateway, possibly through multiple hops.

Of course, in this setting, the reliability of the relay network becomes essential. This means that relay nodes should be suitably deployed in order to achieve a given *fault-tolerance*. Namely, the network configuration should ensure that even if a maximum given number of relay nodes become faulty, all radiogoniometers are still able to deliver their information to the gateway.

Installing a relay node entails two types of costs. First of all, we have to consider the cost of the devices (antennas, amplifiers, etc.). In the second place, and perhaps more importantly, the installation cost must be assessed. The latter may be quite relevant for a critical infrastructure. For instance, installing a relay node in an airport may entail installing antennas in a ground movement area.

The above considerations motivate research on Artificial Intelligence (AI)–based methods to compute placements for relay nodes that *minimise* cost while at the same time *guaranteeing* a given fault tolerance for the network of relay nodes.

## 1.2.  Contributions

In this paper, we present an approach to effectively compute an optimal and fault-tolerant placement of relay nodes in a large critical area.

The main obstacle to be overcome is the huge number of constraints needed to model the problem. In fact, in order to maximise the transmission distance without exceeding WiFi transmission power limits, we deploy directional antennas on relay nodes. In this scenario, radio-visibility between two network nodes should be assessed, therefore we need to represent terrain orography. Since the monitored area is of several square kilometres, many cells should be considered in the discrete grid representing orography for the monitored area.

The proposed algorithm takes as input: the technical (*e.g.*, max transmission power, antenna specifications, etc.) and economical (*e.g.*, cost of devices and installation) characteristics of the relay nodes; the orography of the area to be monitored (*e.g.*, terrain elevation profile, buildings, runways, etc.); constraints on where relay nodes may be installed; desired fault-tolerance $F$.

The generated output is, for each relay node, a routing policy and a placement that minimises the network cost and guarantees that radiogoniometers are connected with the gateway even if $F$ relay nodes become faulty.

In this paper, we illustrate the overall approach describing requirements, solution methods, problem formulation, and experimental results. More specifically, we present:

1. A decision-support tool that, by exploiting specialised algorithms on a High-Performance Computing (HPC) infrastructure, generates (from the input above) a 0/1 Linear Program (0/1-LP) formulation of our optimal relay node placement problem, enabling the use of standard off-the-shelf AI-based reasoners like Mixed Integer Linear Programming (MILP) (see, *e.g.*, [10]), Pseudo-Boolean Satisfiability (PB-SAT) (see, *e.g.*, [11]), as well as Satisfiability Modulo Theory (SMT)/Optimisation Modulo Theory (OMT) solvers (see, *e.g.*, [12]);

2. Experimental results comparing performance of 8 among the current MILP, PB-SAT, as well as SMT/OMT solvers on instances comprising up to 8 million constraints generated from a real case study from the geographic area of the Leonardo da Vinci Airport in Rome (Italy).

Our experimental results show that covering an area of several square kilometres with hundreds of cells generates MILP/PB-SAT/SMT/OMT instances that can be solved within hours of computation on a commodity PC, thus showing the feasibility of the proposed approach.

## 2.  Problem requirements

We consider the problem of enabling reliable and high-throughput communication between radiogoniometer sensors placed throughout the Monitored Area (MA) of an airport to a gateway, typically located at the air traffic control tower. Since the MA is large (with edges that can be several kilometres long) and because of several logistic constraints, a wired communication network is not viable neither technically nor economically, therefore a wireless communication must be used. Moreover, given the long distances involved, and the need, due to regulatory constraints, to use standard WiFi frequencies, *intermediate relay antennas* should be placed to enable multi-hop communication.

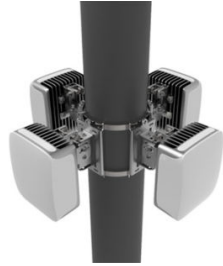

Figure 1: An array of $4$ directional antennas mounted on a pole.

The aim of our *decision support* tool is to compute an *optimal placement* of relay antennas in the MA, in order to design the necessary infrastructure that enables proper routing of the required network traffic from the radiogoniometer sensors (placed near the runways) to the gateway (the air traffic control tower). In this scenario, relay antennas are *directional* antennas, hence they are deployed in *arrays* mounted on poles, as shown in Figure 1. Antennas mounted on the same pole are pointed to different directions. Each antenna array is positioned at a given elevation $L$ from the ground (the height of the poles).

The problem amounts to decide where in the MA and in which direction to deploy relay antennas in order to satisfy all given requirements while minimising the overall cost due to the devices and their deployment. In the following sections we illustrate the high-level problem requirements.

## 2.1.  Monitored Area definition

The MA typically has an irregular shape and a size of several squared kilometres, going from, *e.g.*, the runways to the air traffic control of an airport. Moreover, the terrain orography is irregular: different regions of the MA might show different ground elevations, while obstacles of various types (*e.g.*, buildings like terminals, hangars, etc.) might exist. In this scenario, radio visibility between points in the MA cannot be determined by simply considering their Euclidean distance, because obstacles (*e.g.*, buildings) and the terrain orography (*i.e.*, different elevation) should be considered as well. Therefore, we need to suitably model the *terrain orography*. The orography information about the terrain is stored in available databases. The MA is represented as a partition of *convex polygons*, each associated with a *terrain elevation* value. An example of a MA is provided in Figure 2, in which an airport area is shown with the associated terrain elevation map and obstacles.

## 2.2.  Forbidden placement and link areas

In our scenario, *forbidden placement areas* define those portions of the MA in which no relay node can be placed, while *forbidden link areas* represent those portions of the MA that should not be traversed by radio transmission links. Forbidden placement areas stem from the existence of runways, taxiways, holding areas, etc. in the MA in which no antenna can be installed, as well as other requirements that impose that some regions are kept clear, due to technical or regulatory constraints. Instead, forbidden link areas stem from, *e.g.*, technical or safety requirements (for example, to avoid interference with other radio networks in the airport). Both forbidden placement and forbidden link areas are defined in our database as a set of *convex polygons* lying within the MA. Figure 2 shows forbidden placement and link areas on a sample MA in turquoise and brown, respectively.

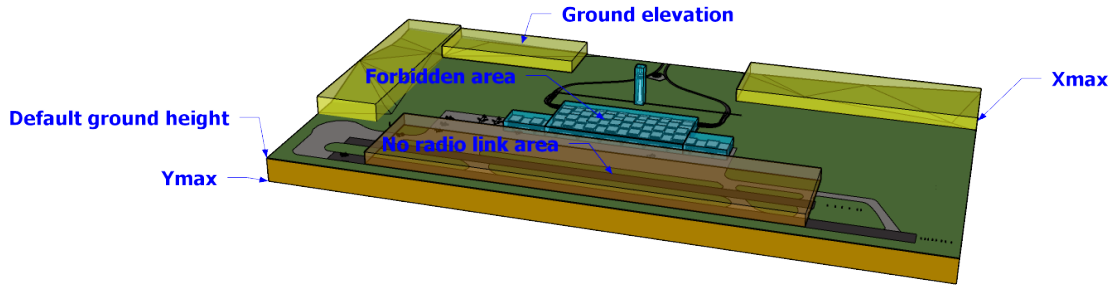

Figure 2: A monitor area that includes an airport, delimited by $X_{max}$, $Y_{max}$, with the associated elevation map and obstacles. The figure highlights the terrain elevation profile (yellow), forbidden placement areas (turquoise), and forbidden link areas (brown).

## 2.3.   Deployment costs

The cost of placing a pole to enable installation of an antenna array is not constant throughout the MA, as it might depend on several parameters (*e.g.*, the accessibility of the chosen point for maintenance). These costs are stored in our database, as a set of *convex polygons* of the MA, each one associated with a value for the cost of deploying a support pole in that area. In contrast, once a support pole has been deployed, mounting a single directional antenna on it has a fixed cost (*i.e.*, the cost for the hardware).

## 2.4.   Radio visibility requirements

Radio visibility requirements state the conditions that must be met by the placement of pairs of directional relay antennas for their network links to work satisfactorily.

In particular, radio communication between two (properly-oriented) antennas in points $(x, y)$ and $(x', y')$ of the MA (at elevation points $h$ and $h'$ as stated by the terrain orography and the height of the holding poles) is considered impossible/unsatisfactory not only if the 3D Euclidean distance between them is longer than a given threshold, but also if the relevant Fresnel Ellipsoid (FE) (see Figure 3) computed for the two antennas intersects obstacles (due to, *e.g.*, terrain elevation changes, buildings, etc.) The Fresnel-zone analysis is indeed needed to ensure limited interference and satisfactory signal quality.

Given two antennas deployed in two points (possibly at different elevations), the relevant FE associated to them is a 3D ellipsoid whose major radius is the straight-line segment connecting the two points, and whose minor radius is the greatest value such that, if a stray component of the transmitted signal bounces off an object within the FE and then arrives at the receiving antenna, the resulting phase shift will be considered to have an unacceptably negative impact on the signal quality. Given the transmitting/receiving power of the employed antennas, their maximum transmission distance and the band used, the value of the minor radius of the relevant FE for the relay placement task at hand can be computed with specialised algorithms. We safely over-approximate the FE model with a bounding box as illustrated in Figure 3.

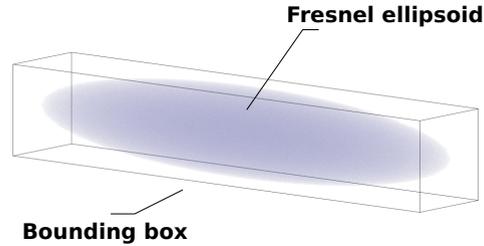

Figure 3: FE defines a 3D region used to assess the strength of radio waves between a transmitter and a receiver. A desired signal quality is associated with an FE zone without obstacles. We safely over-approximate the FE with a bounding box.

## 2.5. Fault tolerance requirements

Fault tolerance requirements ask that the network to be designed guarantees successful routing from each sensor to the gateway also in case of malfunctioning of at most $F \geq 1$ relay nodes. In order to guarantee that fault tolerance requirements are satisfied, the computed relay node placement is required to be such that at least $F + 1$ *distinct routes* from each sensor to the gateway exist, and that two distinct routes from the same sensor *do not share* any relay node.

## 2.6. Performance requirements

Performance requirements ask that the network to be designed ensures that communication from each sensor to the gateway is not significantly impacted by degradation effects, which arise when using too many relay nodes. To guarantee that performance requirements are met, it is requested that each sensor-to-gateway route in the designed network consists of at most $H \geq 1$ hops (*i.e.*, traverses at most $H - 1$ intermediate relay nodes).

## 2.7. Capacity requirements

Network capacity constraints ask that the network is able to convey the needed traffic from all the sensors to the gateway. This means that there is an upper bound to the number of messages (from sensor nodes) that can be routed through a given radio link (*i.e.*, a pair of antennas mounted on different poles each one lying in the radio visibility cone of the other) per second. This limit depends on the transmission band available on radio links and on the band needed to transmit a single message from a sensor.

# 3. Computing optimal fault-tolerant relay placements

Given the intricate requirements outlined in Section 2, a possible approach to solve our problem is to develop a specialised ad-hoc algorithm. However, such an approach would be costly and error-prone, and proving optimality of the computed solution would be not easy.

In this paper we show that a different approach is possible, which exploits standard off-the-shelf AI reasoners to generate *correct-by-construction* intermediate and final results. By lifting-up the abstraction level of the code to be written (when compared to traditional programming), our approach aims at substantially reducing the cost of designing and developing the associated software.

Our algorithm, described in the next sections, is in three steps.

*First*, starting from an input scenario (as outlined in Section 3.1), our algorithm performs a suitable discretisation of the MA (Section 3.2) and populates a relational database with intermediate data that will be needed in the subsequent steps (Section 3.3).

*Second*, a parallel computationally intensive pre-processing step is performed, in order to compute the *radio visibility graph* (Section 3.4). Indeed, requirements related to the terrain orography and the presence of obstacles and radio-visibility conditions between pairs of relay antennas demand for a complex processing of the source data.

*Third*, an optimisation problem is automatically generated and solved (Section 3.5).

All the above steps *extensively use standard AI reasoners to produce correct-by-construction outputs*. In particular, the first two steps define multiple helper computational geometry problems similar to those addressed in [13], and solve them using a standard MILP solver. As for the third step (where we define our main optimisation problem), we experiment with AI reasoners based on different paradigms, namely MILP, PB-SAT, and SMT/OMT.

## 3.1. Input data

From external sources, the following data are available for the scenarios at hand:

1. The geometry and terrain orography of the MA, in terms of a set of convex polygons, each one associated with an elevation value (in metres).

2. The set of areas in the MA where no relay node can be placed (forbidden placement areas), in terms of a set of convex polygons.

3. The set of areas in the MA that cannot be traversed by radio links (forbidden link areas), in terms of a set of convex polygons.

4. A map defining the placement cost of a relay node throughout the MA, again in terms of a set of convex polygons, each one associated with a cost value.

5. The number and positions of the sensor nodes (radiogoniometers) in the MA.

6. The position of the gateway node in the MA.

Moreover, our algorithm takes the following additional inputs:

7.  The maximum transmission rate of relay antennas (all equal).

8.  The transmission rate of each sensor.

9.  The maximum distance $D$ between two relay nodes for their direct radio link to be reliable.

10. The visibility cone aperture $\omega$ of all relay antennas.

11. The value $r$ of the minor radius of the FE to be considered when performing quality assessments of the potential radio links.

12. The maximum number of allowed hops $H$ for all sensors-to-gateway routes.

13. The maximum number of relay node faults $F$ that the network must tolerate.

14. The maximum number of antennas that can be installed in a relay node.

15. The finite set of possible orientations $O$ of antennas in a relay node, in terms of a set of values in degrees from a fixed direction (*e.g.*, North).

16. The fixed cost of installing a single antenna in a (already existing) relay node.

Finally, in order to make the set of candidate relay antenna placements finite, our algorithm considers the MA partitioned in rectangular cells. Each relay node will be placed in a distinct cell. The user is required to provide the desired number of such cells, in terms of:

17. $I$ and $J$ (positive integers), the number of cells in which the area $A$ must be split along the X and Y axes, respectively.

## 3.2.    Monitored area discretisation

The geometry of the MA is over-approximated by a rectangular bounding box $A$. A suitable Cartesian orthogonal coordinate system is defined as to make the edges of rectangle $A$ parallel to the X and Y axes. As a result of this coordinate system transformation, area $A$ is defined by points having X and Y coordinates ranging within $[0, X_{\max}]$ and $[0, Y_{\max}]$ respectively, for some $X_{\max}, Y_{\max} \in \mathbb{R}_+$. From now on, we assume that whenever we need to use geographic coordinates, they will be first transformed accordingly into coordinates of the above system. Any region in $A$ not belonging to the MA will be managed by defining *forbidden placement* (akin to *non-admissible scenarios* in [14]) and link areas in $A$ (see Section 2.2).

Area $A$ is then partitioned into $I \times J$ cells, each one being a rectangle having size $W_X$ by $W_Y$ (in metres), where $W_X = \frac{X_{\max}}{I}$ and $W_Y = \frac{Y_{\max}}{J}$ (see Figure 4).

Cells are denoted by $C_{i,j}$, for $i \in [0, I-1]$ and $j \in [0, J-1]$. Hence, for each $i, j$, $C_{i,j}$ defines the rectangle consisting of points:

$$C_{i,j} = \{(x,y) \in A \mid iW_X \le x \le (i+1)W_X, \; jW_Y \le y \le (j+1)W_Y\}.$$

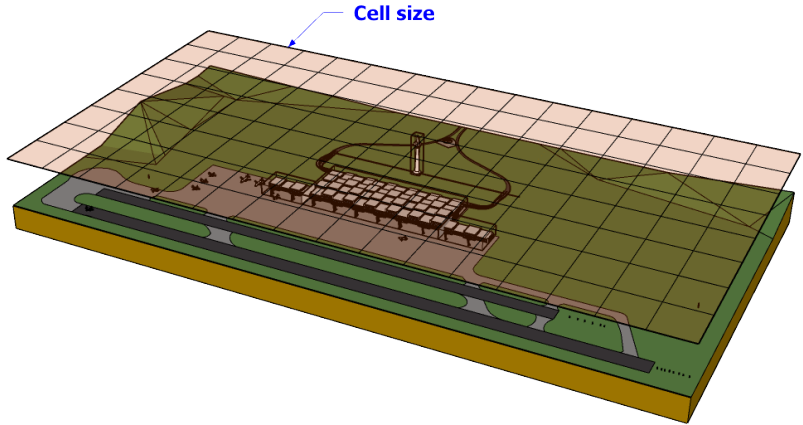

Figure 4: The MA is over-approximated by a rectangular bounding box $A$, which is partitioned into $I \times J$ rectangular *cells*.

## 3.3. Relational database population

The first step of our algorithm is to run helper feasibility and optimisation MILP problems in order to populate a relational database, starting from the input data about terrain orography and pole placement costs. This pre-processing follows the same strategy used in [15] to generate admissible (simulation) scenarios through a pre-processing, basically computing all feasible solutions to a constraint problem.

We proceed as follows. For each cell $C_{i,j} \in A$ we first compute whether it intersects a forbidden antenna placement area. Although this sub-problem is clearly polynomial, we solve it by defining a simple feasibility MILP problem over the linear constraints stemming from the cell definition and those stemming from each of the forbidden placement area convex polygons. The MILP problem also defines one 0/1 decision variable per such polygon. A feasible solution is enforced to set at least one of the 0/1 variables to 1, which in turn imposes that at least one forbidden placement area polygon intersects the cell at hand.

MILP problems as the one outlined above are *trivial* to solve, and define our general correct-by-design approach to check whether a polygon (or, in some cases, a 3D polyhedron) intersects a union of polygons (3D polyhedra). This is done using the same MILP-based approach used in [16] to generate control abstractions. As shown there, this declarative approach greatly simplifies coding and testing, and its impact on computation time is negligible.

For cells $C_{i,j}$ not intersecting a forbidden placement area (hence, whose associated MILP problem is *infeasible*), we also compute: (i) The minimum elevation of a point in $C_{i,j}$; (ii) The maximum pole placement cost within $C_{i,j}$. For such computations, two additional (again trivial) optimisation MILP problems are generated and solved. Constraints of such problems are similar to those of the feasibility MILP problem described earlier to detect intersection between a convex polygon and the union of convex polygons. In this case, also an objective function is defined, which searches for the intersection point having minimum elevation or maximum cost respectively.

At the end of this pre-processing step, much as in [17], our relational database is populated with relation $Cell$, whose tuples $(i, j, e, c)$ define the cells $C_{i,j}$ covering $A$ not intersecting a forbidden placement area together with their minimum elevation value $e$ and maximum antenna placement cost $c$ (positive reals).

## 3.4.    Radio visibility graph computation

The next step performed by our algorithm is to compute the radio visibility graph $\mathcal{G}$ of $A$ using an approach similar to that used in [18, 19] to generate sets of simulation scenarios. The radio visibility graph is an undirected graph $\mathcal{G} = (V, E)$ where:

- The set of nodes $V$ is the set of possible placements of poles within cells of $A$ (*pole nodes*) and possible installations of directional antennas on such poles (*antenna nodes*).

- The set of directed edges $E$ is the set of feasible and reliable communication links that could be established between two antenna nodes installed on different poles and the communication links between each pole and antennas installed on it.

In order to compute the radio visibility graph $\mathcal{G}$, a set of helper feasibility/optimisation MILP problems are solved using data from the input sources.

Sections 3.4.1 and 3.4.2 give more details on the computations of, respectively, nodes and edges of the radio visibility graph, while Section 3.4.3 shows how we parallelised our computation on a large HPC infrastructure.

### 3.4.1.    Computation of nodes

The set of nodes $V$ of $\mathcal{G}$ is computed similarly to scenarios of [20]. In particular, set $V$ is the union of disjoint sets $V_{\text{pole}}$ and $V_{\text{ant}}$, which represent, respectively, *pole* and *antenna nodes*.

Let $O$ be the finite set of possible *orientations* of antennas installed on a pole. For each cell $C_{i,j}$ already proved not to intersect a forbidden placement area (hence such that the pair $(i, j)$ occurs in relation $Cell$ of our database, Section 3.3), $|O| + 1$ nodes are generated:

- A *pole node* of the form $p = (i, j) \in V_{\text{pole}}$, representing the installation of a pole in that cell. For each pole node, the maximum cost for installing a pole within its cell is available in relation $Cell$ of our database (Section 3.3).

- $|O|$ *antenna nodes* of the form $a = (i, j, o) \in V_{\text{ant}}$, one for any possible antenna orientation $o \in O$. The cost of installing a single antenna on an already deployed pole is constant and given as input (Section 3.1).

### 3.4.2.    Computation of edges

Each edge defined in the radio visibility graph $\mathcal{G} = (V, E)$ will represent a possible communication link. The set of edges $E$ is the union of two disjoint sets: $E_{\text{wired}}$ and $E_{\text{radio}}$. Edges in $E_{\text{wired}}$ connects each antenna node to its pole and represents wired connections between any antenna and its pole, and thus, indirectly, with any other antenna mounted on the same pole.

$$E_{\text{wired}} = \left\{ (a,p) \;\middle|\; \begin{array}{l} a = (i,j,o) \in V_{\text{ant}}, p = (i,j) \in V_{\text{pole}}, \\ i \in [0, I-1], j \in [0, J-1], o \in O \end{array} \right\}.$$

Conversely, edges in $E_{\text{radio}}$ represent the possibility of radio links between two antennas mounted on different poles.

$$E_{\text{radio}} \subseteq \left\{ (a,a') \;\middle|\; \begin{array}{l} a = (i,j,o) \in V_{\text{ant}}, a' = (i',j',o') \in V_{\text{ant}}, \\ i, i' \in [0, I-1], j, j' \in [0, J-1], o, o' \in O, (i,j) \neq (i',j') \end{array} \right\}.$$

While computation of $E_{\text{wired}}$ is trivial (as all wirings are possible between an antenna and its pole), computation of $E_{\text{radio}}$ requires to decide, for each pair of antenna nodes in different cells, whether there would be good radio visibility between them, considering their respective orientations, distance, terrain orography, obstacles, and forbidden link areas. To this end, for each pair of antenna nodes $a = (i,j,o)$ and $a' = (i',j',o')$ in different cells (*i.e.*, $(i,j) \neq (i',j')$), our algorithm calls function radio_visibility() (see Algorithm 1) to check whether radio visibility between $(i,j,o)$ and $(i',j',o')$ is satisfactory or not. In the positive case, the algorithm adds both $(a,a')$ and $(a',a)$ to $E_{\text{radio}}$.

Function radio_visibility() first computes the centre positions of the given cells and the minimum heights of the antennas that could be placed on it, and then returns true if and only if: (i) the distance between the two positions is not greater than the maximum transmission distance $D$; (ii) the two antennas (given their orientations) are visible to each other; (iii) no interference is expected between the two antennas due to terrain orography and obstacles, and no forbidden link area is traversed by the prospective radio link.

Double visibility between the two (oriented) antennas is delegated to function double_visibility() that uses standard trigonometry-based computations to check if the second antenna is within the visibility cone of the first one and vice-versa. As antennas, even though directional, might have a visibility cone having non-negligible aperture (value $\omega$ in function double_visibility() of Algorithm 1), parallel radio links can be established between distinct antennas belonging to the same pair of poles.

Computation of absence of interference is delegated to function radio_no_interference() which also takes into account the height of the holding poles $L$. Along the lines of constraint-based over-approximation approaches in [21, 22, 23, 24], function radio_no_interference() computes the 3D bounding box of the FE with minor radius $r$ in terms of linear constraints (hence a safe over-approximation of it), and solves additional helper feasibility MILP problems (along the lines of those outlined in Section 3.3) to check whether such a bounding box intersects at least one polyhedron defined from the terrain orography data (which include obstacles such as buildings) or whether its $XY$ projection overlaps a forbidden link area polygon.

### 3.4.3. Parallel HPC-based computation of the visibility graph

Given that the tasks of deciding whether there is good radio visibility between antenna nodes placed in different cells are independent of each other, and that the number of such such tasks can be huge, our algorithm has been designed in order to exploit a parallel HPC infrastructure for such computations.

---

**Algorithm 1:** Checking feasibility of a radio link between to antennas.

**1 function** radio_visibility( $(i, j, o)$, $(i', j', o')$ )

**2**     x $\leftarrow W_X * (i + \frac{1}{2})$;

**3**     y $\leftarrow W_Y * (i + \frac{1}{2})$;

**4**     h $\leftarrow$ min. elevation of cell $C_{i,j}$ as stored in relation *Cell*;

**5**     x' $\leftarrow W_X * (i' + \frac{1}{2})$;

**6**     y' $\leftarrow W_Y * (i' + \frac{1}{2})$;

**7**     h' $\leftarrow$ min. elevation of cell $C_{i',j'}$ as stored in relation *Cell*;

**8**     **return** $\sqrt{(x' - x)^2 + (y' - y)^2 + (h' - h)^2} \leq D$ *and*

**9**         double_visibility( $(x, y, o)$, $(x', y', o')$ ) *and*

**10**         radio_no_interference( $(x, y, h + L)$, $(x', y', h' + L)$ )

**11 end**

**12 function** double_visibility( $(x, y, o)$, $(x', y', o')$ )

**13**     $\alpha \leftarrow$ vector_angle( $(x', y') - (x, y)$ );

**14**     $\beta \leftarrow$ vector_angle( $(x, y) - (x', y')$ );

**15**     **return** $o - \frac{\omega}{2} \leq \alpha \leq o + \frac{\omega}{2}$ *and* $o' - \frac{\omega}{2} \leq \beta \leq o' + \frac{\omega}{2}$;

**16 end**

---

In particular, our algorithm consists of an Orchestrator and several Worker processes working in parallel on different networked computational nodes. The Orchestrator is in charge of governing the generation of the radio visibility graph. To this end, it computes the set of pole and antenna nodes and the set of antenna-to-its-pole edges (*i.e.*, set $E_{\text{wired}}$). Then, for each pair of antenna nodes in different cells, the Orchestrator delegates to any available Worker the task of executing function radio_visibility() on them, to determine whether the associated edge in the radio visibility graph should be generated or not (*i.e.*, to decide whether this pair of antenna nodes belongs to $E_{\text{radio}}$). In order to reduce the communication overhead, the Orchestrator assigns such edge-computation tasks to Workers in *bunches*. Workload among Workers can be kept balanced by adaptively revising the size of such bunches of tasks, along the lines of [25]. However, as we will experimentally show in forthcoming Section 4.3, for this particular problem the computation time of each task is basically *constant*, since it requires to solve a fixed number of trivial helper MILP problems. This implies that the time needed for each edge-computation task is by far dominated by the time to *start* the helper MILP solver (rather than the time to actually solve such problems, which is negligible).

Hence, the best approach for the Orchestrator to efficiently use the underlying HPC infrastructure is to split *upfront* the set of edge-computation tasks to be performed among the available Workers, *avoiding altogether* the overhead due to (useless) load-balancing. If Workers are deployed on identical machines (which is a typical setting in HPC infrastructures as ours), this approach results in simply splitting upfront the set of edge-computation tasks *evenly* among workers. Our algorithm follows this approach.

Figure 5 shows a sample radio visibility graph as computed by our parallel instance-generation algorithm.

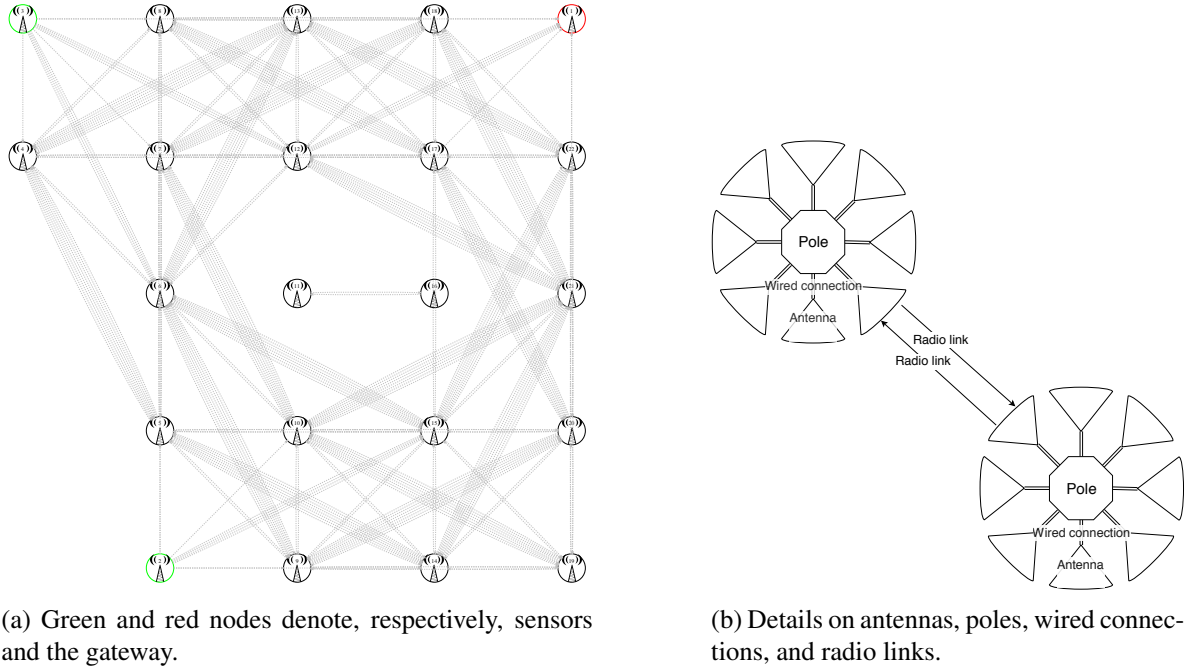

(a) Green and red nodes denote, respectively, sensors and the gateway.

(b) Details on antennas, poles, wired connections, and radio links.

Figure 5: Sample visibility graph.

## 3.5. The main optimisation problem

Thanks to the preprocessing outlined in the previous sections, our radio visibility graph already provides a suitable representation of requirements about forbidden placement and link areas (Section 2.2) and radio visibility (Section 2.4).

Our main optimisation problem described next will decide where to deploy antenna arrays (over holding poles placed in the centre of some of the cells defined as pole nodes in the radio visibility graph) and the number and orientations of the associated antennas, enforcing the remaining requirements (*i.e.*, fault tolerance, performance and capacity requirements, Sections 2.5 to 2.7), while minimising overall deployment costs (as asked by the requirements in Section 2.3).

Given the preprocessing computation of the radio visibility graph described above, the requirements not already handled can be quite conveniently defined as linear constraints. Specifically, we introduce 0/1 decision variables to represent poles, sensors, antennas, gateway, links, and then characterise their relations as linear constraints among these variables. Notice also that such 0/1 variables can be immediately converted into Boolean variables for a PB-SAT specification. This way, the resulting problem specification is a 0/1 Linear Program (0/1-LP) which is suitable for all our solving technologies: MILP, PB-SAT, and SMT/OMT. This uniform problem formulation enables us to compare the performance of these three types of solvers on our industry-scale case study (see Section 4).

Below we outline our main modelling ideas, in terms of a high-level view of decision variables (Section 3.5.1), constraints (Section 3.5.2), and objective function (Section 3.5.3). Full details can be found in our software public repository (see Section 7).

### 3.5.1.    Variables

For each sensor $s$ and for each of the $F + 1$ distinct routes $r$ from $s$ to the gateway (as required by the fault tolerance constraints), the following set of 0/1 decision variables is defined:

$$\{\text{link}_{s,r,t,e} \mid t \in [1, H], e \in E_{\text{radio}}\}$$

where $E_{\text{radio}}$ is the set of *directed* radio links (between antenna nodes in different cells) considered *reliable*. In a solution, assignment 1 to variable $\text{link}_{s,r,t,e}$ denotes that the $r$-th route from sensor $s$ to the gateway uses the (directed) radio link $e$ for its $t$-th hop.

Additional (redundant) 0/1 decision variables are defined to ease the specification of some of the constraints. For example, additional variables are introduced to denote the set of poles and antennas that need to be deployed (their values functionally depend on the assignment to the link variables defined above). Suitable *channelling constraints* (similarly to those described in [26, 27]) are added to the problem specification to keep the values of such variables in sync with those of the variables on which they depend.

### 3.5.2.    Constraints

Given the variables introduced above, the optimisation problem is characterised by several constraints which can be classified in the following types:

*Routes connectedness constraints.* For each sensor $s \in S$, its $F + 1$ required routes to the gateway are made of a sequence of radio links. Connectedness constraints also enforce the use of the proper antenna-to-pole and pole-to-antenna edges for routes traversing two antennas installed on the same pole (such explicit links are needed to enforce capacity constraints).

*Fault tolerance constraints.* For each sensor $s \in S$, its $F + 1$ required routes to the gateway do not share any relay node.

*Number of antennas per relay node.* The antennas to be deployed on each pole do not exceed the upper limit given as input.

*Number of antennas at the gateway.* The antennas to be deployed at the gateway do not exceed the upper limit given as input.

*Network flow and capacity constraints.* A set of constraints (along the lines of [28]) is needed to represent the flow in the graph induced by the computed set of routes. They ensure that each relay node has sufficient transmission capacity to handle the overall network traffic passing through it and that the necessary flow constraints are satisfied.

### 3.5.3.    Objective function

The (linear) objective function asks to minimise the overall placement cost of the relay nodes, which is given by the sum of the cost of placing each pole (these costs depend on the cell where each pole is deployed and have been computed during pre-processing), plus the cost of installing all the required antennas on the deployed poles (the cost of installing each antenna is fixed).

As a variation, we also considered a *feasibility* version of our problem formulation, where the objective function is converted into an additional constraint which enforces that the cost of the found relay node positioning is not above a given threshold (*budget constraint*).

# 4. Experiments

In this section we present our experimental results, focusing on the different phases of our workflow: parallel HPC-based generation of the 0/1-LP problem and its resolution by means of standard off-the-shelf MILP, PB-SAT as well as SMT/OMT solvers. We generated all the 0/1-LP problem instances considered in our experiments using a software we developed in the C language, which is available online (see Section 7).

The outline of this section is as follows. Section 4.1 introduces our case study, based on scenarios from the Leonardo da Vinci airport in Rome, Italy. Section 4.2 briefly outlines our computing infrastructure and lists the 8 MILP, PB-SAT, and SMT/OMT solvers we used to solve our generated 0/1-LP relay node positioning problems. In Section 4.3 we assess the *scalability* of our parallel HPC-based 0/1-LP problem generation, while in Section 4.4 we cope with its actual resolution by means of our 8 solvers. In particular, these sections show that the entire workflow (*i.e.*, 0/1-LP problem generation and resolution) can be completed in *a few hours* on a small computational infrastructure, even when the 0/1-LP problem instances comprise *millions of constraints*.

Note that, for the actual deployment of the relay nodes in a critical infrastructure such as an airport, typical time schedules involve weeks of work. Hence, a few hours of computation can be considered a *negligible time* for our application. This demonstrates the viability of our overall approach, which leverages off-the-shelf AI reasoners to tackle our problem rather than requiring lots of ad-hoc programming.

## 4.1. Case study

We considered several relay-node placement-scenarios for various portions of the Leonardo da Vinci airport (Rome, Italy) area.

### 4.1.1. Scenarios

Table 1 describes the 16 scenarios we considered in terms of:

1. Size of the MA (columns "$X_{max}$" and "$Y_{max}$", in metres);

2. Maximum transmission distance $D$ between two pairs of antennas in meters (column "max tx dist.", in metres);

3. Maximum transmission rate of relay antennas (column "max tx rate", in messages/second);

4. Maximum number $H$ of hops of each sensor-to-gateway route (column "#hops");

| scenario | $X_{max}$ [m] | $Y_{max}$ [m] | max tx dist. [m] | max tx rate [#msg/s] | #hops | max ant/node | FE radius [m] |
|----------|-----------|-----------|------------------|----------------------|-------|--------------|---------------|
| 1  | 100  | 100  | 36   | 5 | 5 | 6 | 3 |
| 2  | 100  | 100  | 36   | 5 | 5 | 6 | 3 |
| 3  | 100  | 100  | 36   | 5 | 5 | 6 | 3 |
| 4  | 3660 | 1270 | 400  | 4 | 5 | 4 | 4 |
| 5  | 3660 | 1270 | 400  | 4 | 5 | 4 | 4 |
| 6  | 3660 | 1270 | 1000 | 4 | 5 | 4 | 4 |
| 7  | 3590 | 1560 | 400  | 4 | 5 | 4 | 4 |
| 8  | 4000 | 3600 | 1000 | 4 | 5 | 6 | 4 |
| 9  | 6040 | 8191 | 1000 | 5 | 5 | 6 | 4 |
| 10 | 6040 | 8191 | 1000 | 5 | 5 | 6 | 4 |
| 11 | 6040 | 8191 | 1000 | 5 | 5 | 6 | 4 |
| 12 | 6040 | 8191 | 1200 | 5 | 5 | 6 | 4 |
| 13 | 6040 | 8191 | 100  | 4 | 3 | 3 | 4 |
| 14 | 6040 | 8191 | 100  | 4 | 3 | 3 | 4 |
| 15 | 6040 | 8191 | 100  | 4 | 3 | 3 | 4 |
| 16 | 6040 | 8191 | 120  | 4 | 3 | 3 | 4 |

Table 1: Leonardo da Vinci Airport scenarios used in our experiments.

5. Maximum number of antennas per node (sensor, relay or gateway, column "max ant/node"). Antennas can be oriented according to the following 8 directions: $O = \{$N, NE, E, SE, S, SW, W, NW$\}$;

6. Minor radius $r$ of the FE (column "FE radius", in metres).

The 16 scenarios differ in terms of the terrain orography, costs for pole and antenna deployments, and technical specifications of the employed relay antennas.

### 4.1.2. Instances

From each of the 16 scenarios in Table 1 we generated multiple instances of our 0/1-LP problem (as an *optimisation* problem), by varying the maximum number $F \in [0, 2]$ of faulty relay antennas that the network must tolerate and the size of the (always square) cells in meters (MA discretisation, see Section 3.2).

As for the latter point, for each scenario in Table 1, we considered square cells having edges of length $W_X = W_Y = \rho W_{tx}$, where:

- $W_{tx} = $ max tx dist./$\sqrt{2}$ is the length of the diagonal of a squared cell having edges long "max tx dist." metres (from Table 1). This is the *largest* (square) cell size that needs to be considered given the sensors at hand (since, with larger square cells, all pairs of nodes would simply be too far for a reliable radio link to be established);

- $\rho \in \{40\% + j10\% \mid j \in [0, 6]\}$ is a reduction factor which defines cells of size varying from 40% to 100% of the above maximum (square) cell size.

As a result, from the 16 scenarios in Table 1 we generated $16 \times 3 \times 7 = 336$ instances of our optimal relay node positioning problem (*i.e.*, aiming to find the positioning having *minimum* cost).

Moreover, we generated the same amount of instances as instances of a *feasibility* problem, aiming at finding any positioning whose cost is not more than a given budget. In our experiments, for each instance such a budget was set to 125% of the minimum positioning cost (as computed when solving the associated optimisation problem instance).

The generated instances comprise 309 to about 6 million zero-one variables and 378 to about 8 million constraints. Such instances are expected to become harder to solve when increasing $F$ and decreasing $\rho$.

## 4.2.   Experimental setting

### 4.2.1.   Computational infrastructure

For our experiments we used 4 networked identical machines, each one equipped with 24GB of RAM, 2 8-virtual-core Intel(R) Xeon(R) CPUs at 2.27 GHz, for an overall 64 cores, and run a Debian 4.9.30-2+deb9u3 (2017-08-06) Linux environment.

The generation of instances from each scenario (which includes the computation of the radio visibility graphs, see Section 4.3) has been performed in parallel using a varying number of cores (from 1 to 64), in order to assess scalability of our parallel 0/1-LP problem generator (see Section 4.3).

Conversely, each such problem instance has been solved (see Section 4.4) with each solver among those described in Section 4.2.2 running separately on a machine, and free to use any number of cores and amount of RAM (within the available quantity). We fixed a *time limit* of 3 hours to all solving processes.

### 4.2.2.   Solvers

We solved our problem instances (both optimisation and feasibility problem instances) using the following 8 state-of-the-art MILP, PB-SAT, and SMT/OMT solvers. All solvers have been used in their default configuration.

**MILP solvers.**   We used the following 3 MILP solvers:

*IBM ILOG CPLEX Optimization Studio V12.8.0*  [29] is a comprehensive solver for optimisation problems (LP, QP, QCP, and MIP).

*GLPK* [30] (GNU Linear Programming Kit) is an open-source package that can be used to solve LP and MILP optimisation problems.

*Gurobi Optimizer* [31] is a commercial solver for LP, QP, QCP, and MIP (MILP, MIQP, and MIQCP) problems.

**PB-SAT solvers.**    We used the following 3 PB-SAT solvers:

*MiniSat+* [32] is based on MiniSat [33]. When first introduced in 2005 during the PB-SAT Evaluation [34], it solved the greatest number of benchmark scenarios.

*NAPS* [35] has found the greatest number of optimum assignments during the 2016 edition of the Pseudo-Boolean Satisfiability Competition [36] in the categories "optimisation, big integers, linear constraints" and "optimisation, small integers, non linear constraints."

*Sat4j* [37] has found a solution (optimal or not) for the greatest number of benchmark scenarios during the 2016 edition of the Pseudo-Boolean Satisfiability Competition [36] in the categories "optimisation, big integers, linear constraints" and "optimisation, small integers, non linear constraints".

**SMT/OMT solvers.**    We used the following 2 SMT/OMT solvers:

*Z3* [38] supports linear optimisation problems through $\nu$Z [39].

*OptiMathSat* [40] has an *inline architecture* which relies on a modified SAT solver to optimise the objective function.

## 4.3.   Instance generation

Here we assess the *scalability* of our parallel approach that, from a scenario (see Section 4.1.1) and suitable values for the cell discretisation reduction factor $\rho$ and the maximum tolerated number of faulty relay antennas $F$ (Section 4.1.2), generates an instance of our relay node placement problem (either as an *optimisation* or a *feasibility* 0/1-LP).

The computation time to generate our instances is *by far* dominated by the time to compute the edges of the radio visibility graph. As explained in Section 3.4, we adopted a parallel approach to compute such a graph, with one Orchestrator assigning to a set of parallel Workers the tasks of deciding whether radio visibility between two antenna nodes mounted on poles placed in different cells is satisfactory or not. Once the radio visibility graph has been computed, the Orchestrator is in charge to dump the encoding of our main optimisation or feasibility 0/1-LP.

In Section 3.4 we argued that, for this particular problem, the best way to efficiently use the underlying parallel computational infrastructure is to split upfront the set of edge-computation tasks among the available Workers (which run on identical machines) in bunches of equal size, thus completely avoiding any communication overhead due to adaptive load balancing. Figure 6 experimentally shows that such a simple approach actually leads to well-balanced edge-computation task bunches. In particular the figure shows the Coefficient of Variation (CV) (*i.e.*, standard deviation divided by average) of the time needed by any given number $k \in \{8, 16, 32, 64\}$ of parallel Workers to complete their activity, when the Orchestrator assigns them equal-sized bunches of edge-computation tasks (one per

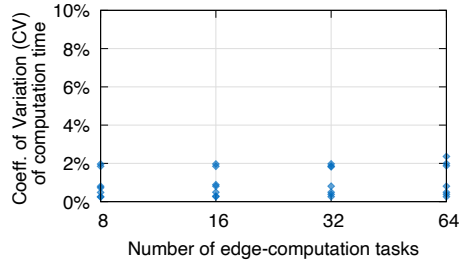

Figure 6: CV of the time to decide radio visibility between two antenna nodes.

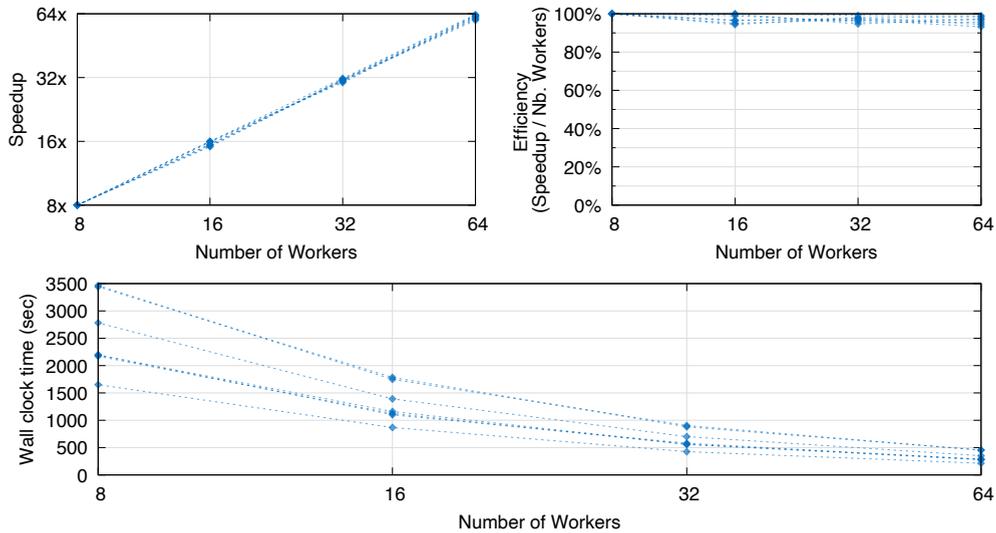

Figure 7: Speed-up, efficiency and overall completion time of our parallel instance generation algorithm, when run on 7 among the largest scenarios.

Worker). The figure shows such statistics on 7 among the largest instance generation processes. We can see that the CV is very stable and always extremely low, thus suggesting that any adaptive load-balancing approach (which would require additional communication overhead) would be simply not worth deploying or might be even deleterious for the overall performance of the algorithm.

Figure 7 shows speed-up, efficiency, and completion time of our overall parallel algorithm when run on the same 7 (among the largest) instance generation processes and using $k \in \{8, 16, 32, 64\}$ Workers. The *extremely high efficiency* achieved by our algorithm (always above 90%) for all values of $k$ further confirms that our simple implementation exploits the underlying parallel computational infrastructure at its maximum. Figure 7 (bottom) finally shows that instances can be generated in a few minutes of computation on a small parallel infrastructure also when starting from the largest scenarios.

## 4.4.    Computing optimal placement of relay nodes

In this section we show the performance of the 8 considered MILP, PB-SAT, and OMT/SMT solvers, when run on both optimisation and feasibility instances of our relay node positioning problem, as generated in Section 4.3.

### 4.4.1.    Optimisation instances

**Solvers scalability.**    Figure 8 shows, for each number $c$ of constraints, the overall number of generated instances having at most $c$ constraints (curve "Overall instances") and the number of such instances successfully solved by each solver within our time limit (3 hours).

From the figure we see that MILP solvers are able to solve the *greatest number* of and the *largest* instances (*i.e.*, instances with up to around 2-million constraints), with CPLEX clearly being the top-performer, followed by Gurobi and GLPK. PB-SAT solvers succeeded to solve instances with up to around half-a-million constraints (with Sat4j being able to solve some of the largest instances, MiniSat+ immediately following, and NAPS lagging behind). OMT solvers were able to solve fewer of the largest instances, with Z3 slightly outperforming OptiMathSat.

Overall, Figure 8 interestingly shows that all solvers were able to solve real-world and hard instances of such an industry-relevant optimisation problem, consisting of a remarkably high number of constraints.

**Comparison of solvers performance.**    Figure 9 compares the performance of our 8 solvers. In particular, for each solver $s$ and each instance having $c$ constraints solved by $s$ in $t$ seconds (within our time limit of 3 hours), Figure 9 shows a dot in position $(c, t)$ whose colour is associated to $s$. Figures 9a and 9b focus on the subsets of satisfiable and unsatisfiable instances, respectively.

The figure shows that MILP solvers typically outperform both PB-SAT and OMT solvers and that PB-SAT solvers usually outperform OMT solvers. Noticeably, the performance of PB-SAT solvers is competitive to those of MILP solvers on unsatisfiable instances.

OMT solvers seem to have very hard time in finding an optimal solution on satisfiable instances. Interestingly, OMT solvers generally show better performance on unsatisfiable instances.

Finally, Figure 10 shows, for each of the 16 scenarios of Table 1 and each value for the MA discretisation factor (in terms of $\rho$, see Section 4.1.2), the maximum value for $F$ (fault tolerance) for which we were able to compute an optimal positioning or to prove unsatisfiability within the 3 hours time-limit. The point associated to each instance is coloured according to the *fastest* solver for that instance. We see that MILP technology is a hands-down winner on the entire spectrum of scenarios, with CPLEX being the fastest solver on almost all instances (with the interesting exception of Gurobi and GLPK, which are faster than CPLEX on a few large instances).

### 4.4.2.    Feasibility instances

In this section we compare the performance of our 8 solvers on *satisfiable feasibility* instances of our relay node positioning problem (*i.e.*, an instance actually *solved* by at least one solver). Such feasibility instances aim at finding *any* relay node positioning whose cost is not above a given threshold

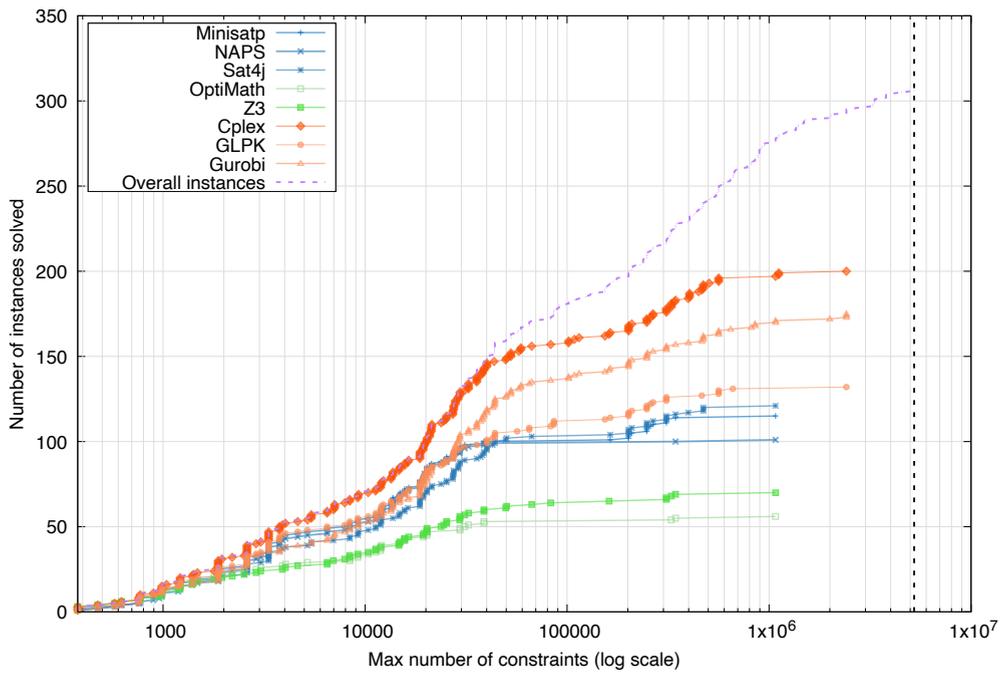

Figure 8: Number of available instances with any given maximum number of constraints solved by each solver within the time limit (3 hours). The vertical dashed line denotes the maximum number of constraints in an instance.

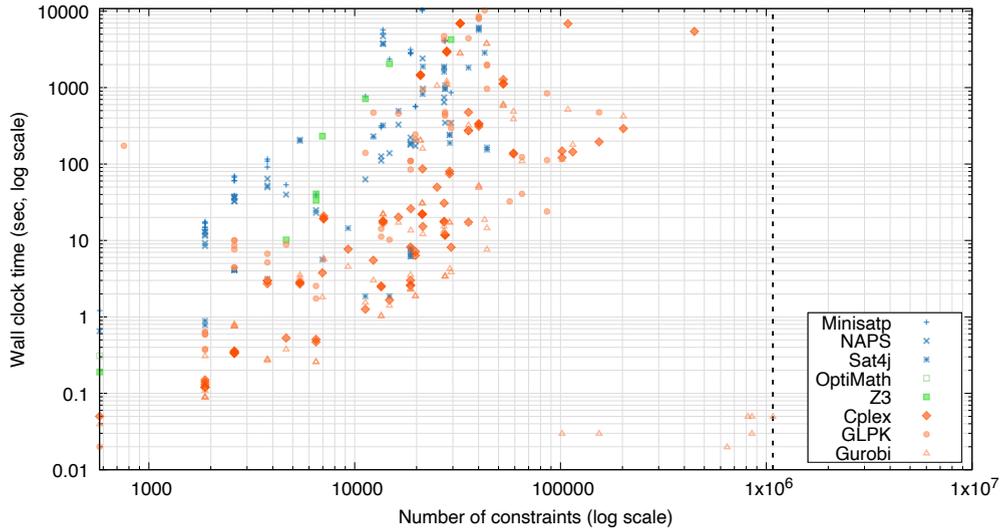

(a) Satisfiable instances (optimal solution found)

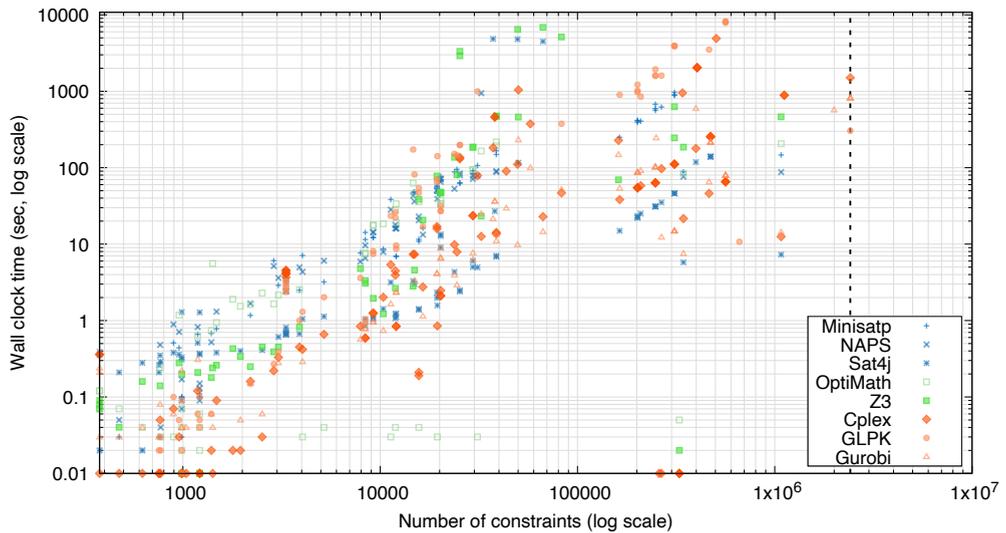

(b) Unsatisfiable instances (unsatisfiability proved)

Figure 9: Comparative performance of our 8 solvers on the generated *optimisation* satisfiable and unsatisfiable instances (time limit: 3 hours). The vertical dashed lines denote the maximum number of constraints in a satisfiable and unsatisfiable instance (*i.e.*, an instance actually *solved* by at least one solver).

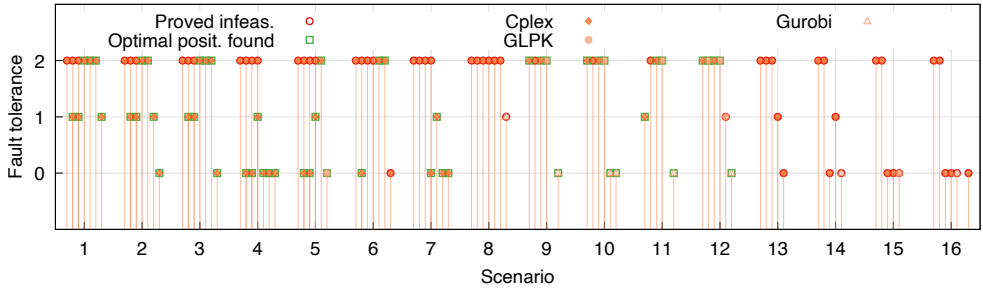

Figure 10: Maximum fault-tolerance optimal relay node positionings computed on our 16 scenarios. For each scenario, we report results for the chosen 7 values of the MA discretisation factor $\rho$, going from 100% to 40% from left to right.

(*budget*). The budget has been set to be 125% of the cost of the optimal positioning, as computed in Section 4.4.1, thus basically introducing a *uniform relaxation* in all satisfiable instances. Our aim is to assess to what extent the absence of an objective function and a problem relaxation makes the problem easier to solve for the different solving technologies.

Figure 11 shows the results of our experiment. In particular, Figure 11a shows, for each number $c$ of constraints, the overall number of generated feasibility instances having at most $c$ constraints (curve "Overall instances") *and* proved *satisfiable* in Section 4.4.1, together with the number of such instances successfully solved by each solver within our time limit (3 hours). Figure 11b compares the performance of our 8 solvers on such instances: for each solver $s$ and each instance having $c$ constraints solved by $s$ in $t$ seconds (within our time limit), Figure 11b shows a dot in position $(c, t)$ whose colour is associated to $s$.

It can be seen that, unsurprisingly, all solvers are in general faster to cope with feasibility (rather than optimisation) instances. Problem relaxation is particularly beneficial to SMT solvers, which are now able to cope with much larger instances (up to around $30\,000$ constraints). As for MILP and PB-SAT solvers however, problem relaxation does not generally allow to cope with significantly larger instances than the optimisation instances solved in Figure 9a.

## 5. Related work

A survey on node placement in wireless sensor networks is presented in [41]. None of the methods discussed therein focus on minimising network cost while at the same time guaranteeing network connectivity notwithstanding loss (*faults*) of up to a given number of relay nodes. We note that random deployment structurally achieves some degree of fault tolerance, but on one side the network cost is far from being minimal and on the other side no (deterministic) guarantee on the number of faults that the network can withstand can be given *a priori*.

An artificial bee colony algorithm for optimal placement of relay nodes in wireless sensor networks has been presented in [42]. The goal of the authors is to reduce the communication holes stemming from randomly-placed relay nodes. Results show that indeed a clever positioning of relay nodes

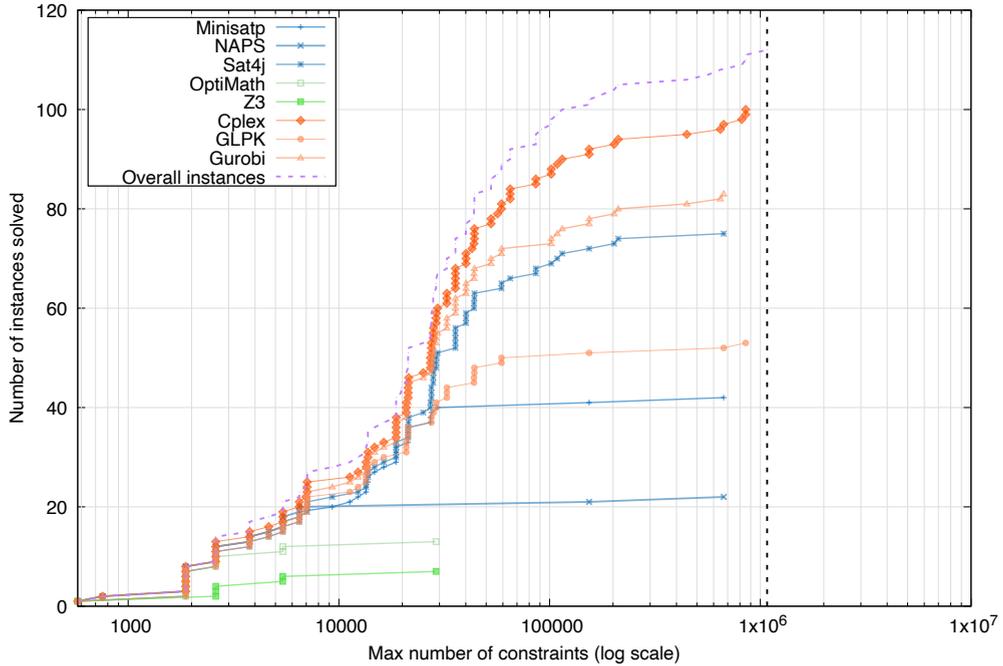

(a) Feasible solutions found

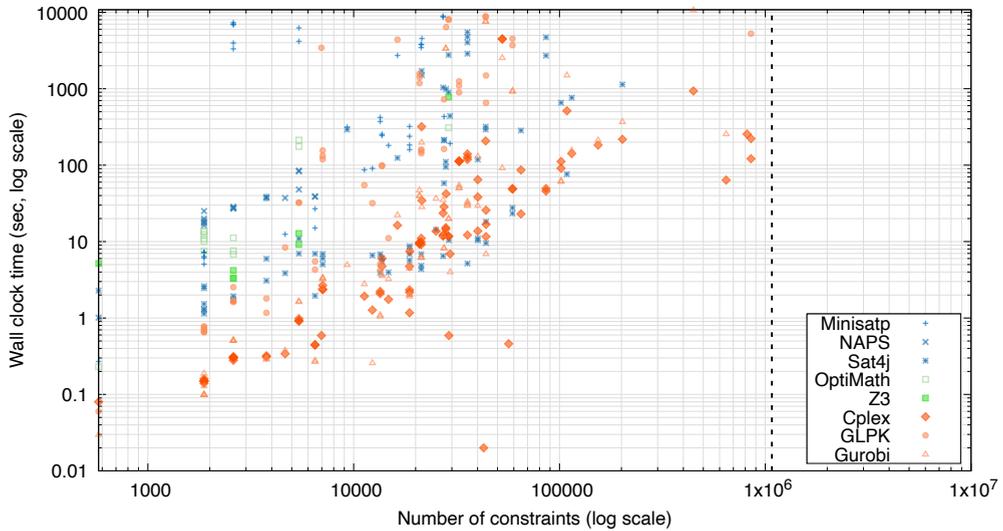

(b) Comparative performance of our 8 solvers.

Figure 11: Scalability and comparative performance of our 8 solvers on the generated *feasibility* satisfiable instances (placement cost at most 125% of cost of optimal placement; time limit: 3 hours). The vertical dashed lines denote the maximum number of constraints in a satisfiable instance (*i.e.*, an instance actually *solved* by at least one solver).

decreases the network energy consumption. Unfortunately, the methods in [42] cannot be directly used in our context since energy is not a concern for us, whereas our goal is to minimise network cost while guaranteeing a given fault tolerance.

The relationship between energy consumption of network relay nodes and routing strategies is studied in [43] as a maximum flow problem and solved partly using a MILP approach and partly using a heuristic approach. We note that in our setting relay nodes do not have energy problems since they are connected to the electrical grid. Conversely, our main problem here is to find a positioning of the relay nodes and antennas with minimal cost. To this end, differently from [43], we need to choose both the minimal number of relay nodes and the optimal orientations for the directional antennas installed on each of them.

In [44] a 2D model for directional antennas is presented. Such a model can also be used in our context, since in our case each relay can host up to 8 directional antennas.

Most of the research work has focused on routing through relay nodes with a known given position (*e.g.*, see [45, 46, 47, 48]). Our scenario is quite different, since we have to compute at the same time the position of the relay nodes as well as the routing strategy for each relay node. Again, differently from other works, our solution also defines the orientation of the antennas hosted by each relay node.

The transmission capacity of ad-hoc networks has been studied in [49]. In our context, such results can be used to model transmission capacity between sensor nodes (*i.e.*, *radiogoniometers*) and gateway nodes.

Modelling of radio communication obstacles in a real environment has been studied in [50]. Such results can be also used in our context to model static transmission obstacles such as walls, hills, etc.

Network traffic, congestion and interference have been analysed in [51] using a MILP-based approach. While such results cannot be directly used in our context, the methods in [51] are useful in our setting to analyse the solution process.

Optimal positioning of relay nodes in a WiFi network (our setting) has been studied in [46]. We note however that [46] does not address fault-tolerant positioning, which is the main problem we have to address in our scenario.

The work in [47] gives up optimality and presents heuristic methods for fault-tolerant positioning of relay nodes. Although we also ask for optimality, some of the considerations in [47] could be exploited also in our case to restrict the search space.

The work in [52] presents methods for optimal and robust (with respect to throughput degradation) positioning of relay nodes. Although in our setting *robustness* in [52] is not the same as fault-tolerance, clearly the work in [52] is aiming towards the same direction as ours.

Summing up, although there are papers addressing optimal positioning of relay nodes as well as papers addressing fault-tolerant positioning of relay nodes, to the best of our knowledge there is no previously published work addressing optimal fault-tolerant positioning of relay nodes.

## 6. Conclusions

We addressed the problem of computing a deployment for relay nodes in a wireless network that minimise the relay node network cost while at the same time guaranteeing proper working of the network even when some of the relay nodes (up to a given maximum number) become faulty (*fault-tolerance*).

For such a problem, we presented an algorithm that, in a few minutes of computation on a small parallel infrastructure, produces a 0/1-LP formulation of either an optimisation or a feasibility problem which can be straightforwardly given as input to different classes of standard AI solvers, in particular: MILP, PB-SAT and SMT/OMT.

Our experimental results using 8 among the state-of-the-art MILP, PB-SAT and SMT/OMT solvers show that we can compute in a few hours (hence, within a fully acceptable time given our application domain) an optimal placement of relay nodes in areas of 40+ squared kilometres (*i.e.*, the area of a medium-large airport) with a precision of a few hundred metres.

In future work we aim at experimenting with additional solving technologies like, *e.g.*, Constraint Programming (see, *e.g.*, [53, 54]) or Constraint Answer Set Programming (CASP, see, *e.g.*, [55, 56, 57]). Also, in order to further increase the size of the areas we can manage, we plan to hybridise our approach with iterative improvement techniques like local-search and meta-heuristic methods (see, *e.g.*, [58, 59, 60]) specially suited for large instances (*e.g.*, those in [61]), of course sacrificing optimality guarantees. Finally, we aim at reducing the need of linear constraint over-approximations by exploiting simulation-based approaches (*e.g.*, [62, 63, 64, 65]) driven by intelligent search (*e.g.*, [66, 67]).

## 7.   Software availability

Our software is publicly available, as a Docker container, at:

https://bitbucket.org/mclab/optimal-relay-node-placement-tool/

The Docker container allows the user to generate the 0/1-LP formulation of our relay node positioning problem from a given input scenario, both as an optimisation and a feasibility problem. The 0/1-LP problems are generated in the languages used by the MILP, PB-SAT, and SMT/OMT solvers included in our experiments.

**Acknowledgements**   This work was partially supported by: Italian Ministry of University and Research under grant "Dipartimenti di eccellenza 2018–2022" of the Department of Computer Science of Sapienza University of Rome; INdAM "GNCS Project 2019".

Authors are grateful to the anonymous reviewers and to A. Scialanca, F. Lanciotti, R. Guarnieri, S. Di Pompeo for kindly providing the data we used in our experimental evaluation and for interesting discussions.